\documentclass[]{spie}  

\usepackage[utf8]{inputenc} 
\usepackage[T1]{fontenc}    
\usepackage{nicefrac}       
\usepackage{mathtools,amssymb,amsthm, amsfonts}
\usepackage{graphicx}
\usepackage[colorlinks=true, allcolors=blue]{hyperref}
\usepackage[nameinlink]{cleveref}
\usepackage{listings}
\hypersetup{
colorlinks=true,
bookmarks=true,
bookmarksdepth=2
}
\usepackage{bookmark}
\usepackage{longtable,booktabs}     
\usepackage{subcaption}
\usepackage[shortlabels,inline]{enumitem}
\setlist{nosep}

\newcommand{\cD}{\mathcal{D}}

\numberwithin{equation}{section}

\theoremstyle{plain}

\theoremstyle{definition}

\theoremstyle{remark}

\DeclareMathOperator{\Acc}{Acc}
\DeclareMathOperator{\PS}{PS}

\newcommand{\mycite}[1]{[\citenum{#1}]}

\title{Impact of architecture on robustness and interpretability of multispectral deep neural networks}

\author[a]{Charles Godfrey}
\author[a]{Elise Bishoff}
\author[a,b]{Myles McKay}
\author[a]{Eleanor Byler}
\affil[a]{Pacific Northwest National Laborotory, Richland, WA, USA}
\affil[b]{University of Washington, Seattle, WA, USA}

\authorinfo{Further author information: send correspondence to godfrey.cw@gmail.com and eleanor.byler@pnnl.gov.}

\begin{document} 
\maketitle

\begin{abstract}
   Including information from additional spectral bands (e.g., near-infrared)
   can improve deep learning model performance for many vision-oriented tasks.
   There are many possible ways to incorporate this additional information into
   a deep learning model, but the optimal fusion strategy has not yet been
   determined and can vary between applications.  At one extreme, known as
   ``early fusion,'' additional bands are stacked as extra channels to obtain an
   input image with more than three channels. At the other extreme, known as
   ``late fusion,'' RGB and non-RGB bands are passed through separate branches of
   a deep learning model and merged immediately before a final classification or
   segmentation layer. 
   
   In this work, we characterize the performance of a suite
   of multispectral deep learning models with different fusion approaches,
   quantify their relative reliance on different input bands and evaluate their
   robustness to naturalistic image corruptions affecting one or more input channels. 
\end{abstract}


\section{INTRODUCTION}

Many datasets of overhead imagery, in particular those collected by satellites,
contain spectral information beyond red, blue and green (RGB) channels (i.e.
visible light). With the development of new sensors and cheaper launch vehicles, the availability of such multispectral overhead images has grown rapidly in the last 5 years. Deep
learning applied to RGB images is at this point a well-established field, but by
comparison, the application of deep learning techniques to multispectral imagery is
in a comparatively nascent state. From an implementation perspective, the
availability of additional spectral bands expands the space of neural network
architecture design choices, as there are a plethora of ways one might ``fuse''
information coming from different input channels. From an evaluation
perspective, multispectral imagery provides a new dimension in which to study model robustness. While in a recent years there has been a large amount of research
on robustness of RGB image models to naturalistic distribution shifts such as
image corruptions, our understanding of the robustness of multispectral neural
networks is still limited. Such an understanding will be quite valuable, as an honest
assessment of the trustworthiness of multispectral deep learning models will
allow for more informed decisions about deployment of these models in
high-stakes applications and use of their predictions in downstream analyses. 

We take a first step in this direction, studying multispectral models operating
on RGB and near-infrared (NIR) channels on two different data sets and tasks
(one involving image classification, the other image segmentation). In addition,
for each dataset/task we consider two different multispectral fusion
architectures, early and late (to be described below).  Our findings
include: 
\begin{enumerate}[(i)]
   \item Even when different fusion architectures achieve near-identical
   performance as measured by test accuracy, they leverage information
   from the various spectral bands to varying degrees: we find that
   for classification models trained on a dataset of RGB+NIR overhead
   images, late fusion models place far more importance on the NIR band in
   their predictions than their early fusion counterparts. 
   \item In contrast, for segmentation models we observe that both fusion styles resulted in models that place greater importance on RGB channels, and
   this effect is \emph{more pronounced} for late fusion models.
   \item Perhaps unsurprisingly, these effects are mirrored in an evaluation of
   model robustness to naturalistic image corruptions affecting one or more input channels --- in particular, early fusion classification models are more
   sensitive to corruptions of RGB inputs, and segmentation models with either
   architecture are comparatively immune to corruptions affecting NIR inputs
   alone.
   \item On the whole, our experiments suggest that segmentation models and
   classification models use multispectral information in different ways.
\end{enumerate}

\section{RELATED WORK}

The perceptual score metric discussed in \cref{sec:ps} was introduced in
\mycite{gat2021perceptual}, and it can be viewed as a member of a broader
family of model evaluation metrics based on ``counterfactual examples.'' For a
(by no means comprehensive) sample of the latter, see
\mycite{goyalCounterfactualVisualExplanations2019a,
changExplainingImageClassifiers2022}, and for some cautionary tales about the
use of certain counterfactual inputs see \mycite{jainMissingnessBiasModel2022a}. 

For a look at the state of the art of machine learning model robustness, we
refer to \mycite{hendrycksUnsolvedProblemsML2022}. The work most directly
related to this paper was the creation of the ImageNet-C dataset
\mycite{hendrycksBenchmarkingNeuralNetwork2019}, obtained from the ImageNet
\mycite{imagenet_cvpr09} validation split by applying a suite of naturalistic
image corruptions at varying levels of severity. The original ImageNet-C paper
\mycite{hendrycksBenchmarkingNeuralNetwork2019} showed that even state-of-the-art image classifiers that approach human accuracy on clean images suffer
severe performance degradation on corrupted images (even those that remain
easily recognizable to humans). More recent work
\mycite{kamannBenchmarkingRobustnessSemantic2020} evaluated the corruption
robustness image \emph{segmentation} models, finding that their robust accuracy
tends to be correlated with clean accuracy and that some architectural
features have a strong impact on robustness. All of the research in this paragraph deals with RGB imagery alone.

There is a limited amount of work on robustness of multispectral models, and
most of the papers we are aware of investigate \emph{adversarial robustness},
i.e. robustness to worst case perturbations of images
\mycite{ortizDefenseAdversarialExamples2018,yuInvestigatingVulnerabilityAdversarial2020,duAdversarialAttacksSatelliteborne2021a}
generated by a hypothetical attacker exploiting the deep learning model in question.
It is worth noting that there is a lively ongoing discussion about the realism
of the often-alleged security threat posed by
adversarial examples \mycite{gilmerMotivatingRulesGame2018}.
The only research we are aware of addressing robustness of multispectral deep
learning models to \emph{naturalistic} distribution shift is
\mycite{podsiadloStudyRobustnessLong2020}, which studies the robustness of
land cover segmentation models evaluated on images with varying level of
occlusion by clouds.


\section{DATASETS, TASKS AND MODEL ARCHITECTURES}

The RarePlanes dataset \mycite{Shermeyer2020RarePlanesSD} includes 253 Maxar
Worldview-3 satellite scenes including \(\approx 15,000\) human annotated
aircraft. Crucially for our purposes this data includes RGB, multispectral and
panchromatic bands. In addition
to bounding boxes identifying the locations of aircraft, annotations contain
meta-data features providing information about each aircraft. Of particular
interest is the \textbf{role} of an aircraft, for which the possible values are
displayed in \cref{tbl:role-feat}.
\begin{longtable}[]{@{}cc@{}}
   \toprule()
   Attribute & Sub-attribute \\
   \midrule()
   \endhead
   Civil & \{Large, Medium, Small\} Transport \\
   Military & Fighter, Bomber,* Transport, Trainer* \\
   \bottomrule()
   \caption{The RarePlanes role meta-data feature. Sub-attributes marked with
   \(\ast \) have fewer than 10 training examples and are omitted from our classification dataset.}\label{tbl:role-feat}
\end{longtable}
 
Using this information, we create an RGB+NIR image classification data set with the
following processing pipeline: beginning with the 8-band, 16-bit multispectral
scenes, we apply pansharpening, rescaling from 16- to 8-bit, contrast
stretching, and gamma correction (for further details see \cref{sec:img-proc}). This results in 8-band, 8-bit scenes, from
which we obtain the RGB and NIR channels. \footnote{All NIR inputs in this work refer to the WorldView NIR2 channel, which covers 860-1040nm.}
To extract individual plane images or "chips"
from the full satellite images, we clip the image around each plane using the
bounding box annotations. We use the plane
role sub-attributes (from the right column of \cref{tbl:role-feat}) as
classification labels, and omit the "Military Trainer" and "Military Bomber" classes, which only have 15 and 6 examples, respectively. The remaining five classes have approximately 14,700 datapoints.

We create a train-test split at the level of full satellite images. Note that
this this presumably results in a more challenging machine learning task
(compared with randomly splitting after creating chips), since it requires a
model to generalize to new geographic locations, azimuth and sun elevation
angles, and weather. We further divide the training images into a training and
validation split, and keep the test images for unseen, hold-out evaluations. The
final data splits are spread 74\%/13\%/13\% between training, validation, and
test, with class examples as evenly distributed as possible. The RarePlanes data
set also includes a large amount of synthetic imagery --- however, the synthetic data only includes RGB imagery. Thus, in our experiments we
only use the real data. 

\begin{figure}[tb]
   \centering
   \begin{subfigure}{0.75\linewidth}
      \centering
      \includegraphics[width=\linewidth]{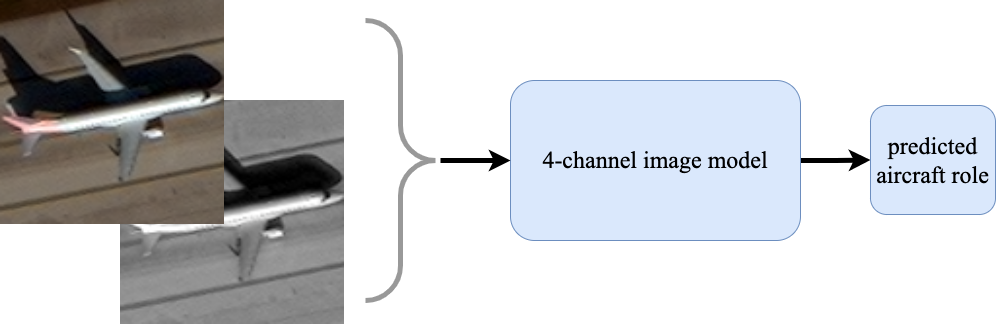}
   \end{subfigure}
   \begin{subfigure}{0.75\linewidth}
      \centering
      \includegraphics[width=\linewidth]{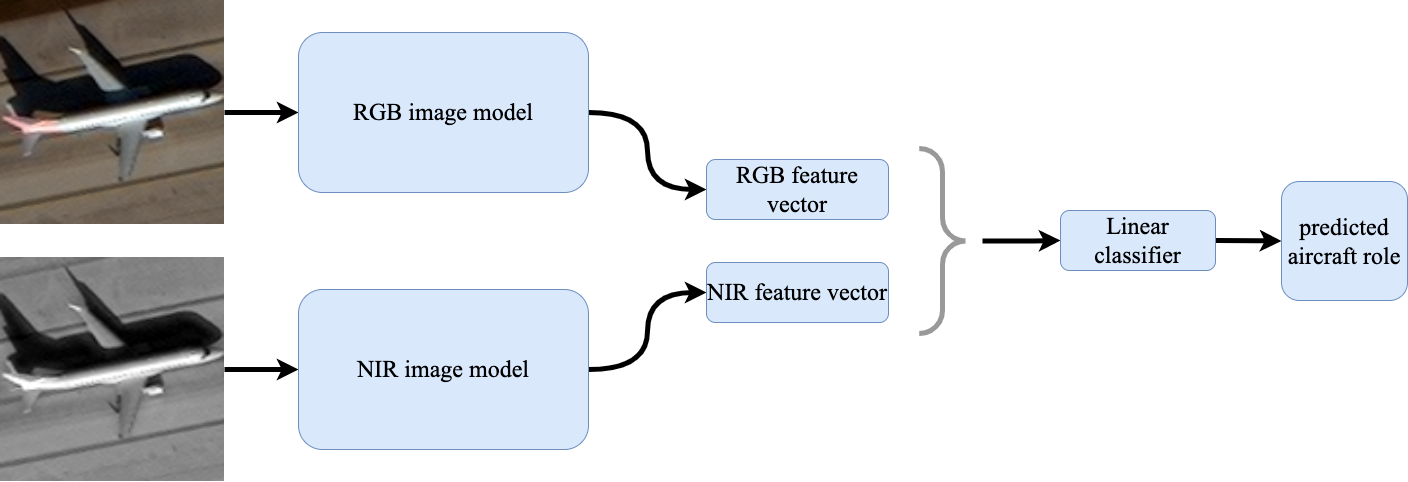}
   \end{subfigure}
   \caption{RGB+NIR fusion architectures. \textbf{Top}: early, \textbf{Bottom}:
   late. Braces denote image/feature concatenation.}\label{fig:fusion-arch}
\end{figure}

We train four different types of image classifiers on the RGB+NIR RarePlanes
chips, all assembled from ResNet backbones \mycite{heDeepResidualLearning2015}:
\begin{description}
   \item[RGB] A standard ResNet34 operating on the RGB channels (NIR is ignored)  
   \item[NIR]  A  ResNet34 operating on the NIR channel alone (RGB is ignored)
   \item[early fusion] RGB and NIR channels are concatenated to create a
   four channel input image, and passed into a 4 channel ResNet34
   \item[late fusion] RGB and NIR channels are passed into \emph{separate}
   ResNet34 models, and the pentultimate hidden feature vectors of the respective
   models are concatenated --- the concatenated feature factor is then passed to a
   final classification head.
\end{description}
The two fusion architectures are illustrated in \cref{fig:fusion-arch}.
We train these RarePlanes image classifiers using \emph{transfer learning}:
rather than beginning with randomly initialized weights, wherever possible we
start with ResNet34 weights pre-trained on the ImageNet dataset
\mycite{imagenet_cvpr09}, and then fine-tune with continued stochastic gradient
descent to minimize cross entropy loss on RarePlanes. Notably, for models with NIR input, the first layer convolution weights are initialized with the Red channel weights trained on ImageNet. Further architecture, initialization and
optimization details can be found in \cref{sec:mod-arch-train}. All model
accuracies lie in the range \(91.8-92.5\% \).

We also use the Urban Semantic 3D  (hereafter US3D)  dataset \mycite{christie2020geocentricpose,christie2021geocentricpose,saux2019,bosch2019} of
overhead 8-band, 16-bit  multispectral images and LiDAR point cloud data with segmentation
labels. The US3D segmentation labels consist of seven total classes, including
ground, foliage, building, water, elevated roadway, and two "unclassified"
classes, corresponding to difficult or bad pixels. From this data we create an
image \emph{segmentation} dataset via the following procedure: First, the
8-band, 16-bit  multispectral images are converted to 8-bit RGB+NIR images with
a pipeline similar to the one used for RarePlanes above. The resulting images
are quite large, and are subdivided into 27,021 \(1024\times1024\)
non-overlapping ``tiles'' with associated segmentation labels. Again,
train/validation/test splits are created at the level of parent satellite
images, and care is taken to ensure that the distributions of certain
meta-data properties (location, view-angle and azimuth angle) are relatively
similar from one split to the next. Further details can be found in
\cref{sec:img-proc}.

Our segmentation models use the DeepLabv3 architecture
\mycite{chenRethinkingAtrousConvolution2017}. To obtain neural networks taking
both RGB and NIR images as inputs, we can apply the strategy 
described in \cref{fig:fusion-arch} to the \emph{backbone} of DeepLabv3 (for a
more detailed description of DeepLabv3's components see
\cref{sec:mod-arch-train}). More precisely, we consider four different
backbones:
\begin{description}
   \item[RGB] A ResNet50 operating on the RGB channels (NIR is ignored)  
   \item[NIR]  A ResNet50 operating on the NIR channel alone (RGB is ignored)
   \item[early fusion] RGB and NIR channels are concatenated to create a
   four channel input image, and passed into a 4 channel ResNet50
   \item[late fusion] RGB and NIR channels are passed into \emph{separate}
   ResNet50 models, and the resulting feature vectors of the respective
   models are concatenated --- the concatenated feature factor is then passed to
   the DeepLabv3 segmentation head.
\end{description}
As in the case of our RarePlanes experiments, we fine-tune pre-trained weights,
beginning with segmentation models pretrained on the COCO datatset
\mycite{linMicrosoftCOCOCommon2014} and again initializing both the first-layer R
and NIR convolution weights with the R channel weights trained on COCO.
In comparison to training the RarePlanes models, this optimization problem is
far more computationally demanding, due to the larger tiles, larger networks and
more challenging learning objective. We used (data parallel) distributed training
on a cluster computer to scale batch size and reduce training and evaluation
time. All model validation IoU scores lie in the range \(0.53-0.55\) --- for details and hyper parameters see \cref{sec:mod-arch-train}.

\section{PERCEPTUAL SCORES OF MULTISPECTRAL MODELS}
\label{sec:ps}

Given a neural network processing multispectral (in our case RGB+NIR) images,
one can ask is which bands the model is leveraging to make its predictions. More
generally, we may want to know the relative importance of each spectral band for
a given model prediction. This information is of potential interest for a number
of reasons:
\begin{itemize}
   \item For many objects of interest, reflectance properties vary widely between spectral bands (for example, plants appear vividly in the NIR band). Depending on the
   machine learning task and underlying data, this phenomenon could cause a
   multispectral model to prioritize one of its input channels.
   \item Some spectral bands are less affected by adverse weather or environmental conditions. For example, NIR light can penetrate haze, and NIR imagery is often used by human analysts to help discern detail in smoky or hazy scenes.
   \item In some applications, the different bands included in a multispectral
   dataset could have been captured by different sensors\footnote{This is not the
   case for our data sets, which were both derived from 8-band images captured
   with a single sensor.}. For example, an autonomous vehicle may be equipped with an RGB camera and a thermal IR camera mounted side-by-side. In such a situation, technical issues affecting one sensor
   could result in image corruptions that only affect a subset of
   channels, and the performance of the downstream model predictions would depend on the relative importance of the corrupted channels.
\end{itemize}

A simple baseline for assessing the relative the importance of input channels
for the predictions of a multispectral model is provided by the
\textbf{perceptual score} metric \mycite{gat2021perceptual}, computed for RGB and
NIR channels as follows: let \(f(x_{\text{RGB}}, x_{\text{NIR}}) \) be an
RGB+NIR model, where \(x_{\text{RGB}}\) and \( x_{\text{NIR}} \) denote the RGB
and NIR inputs respectively. Let 
\begin{equation}
   \cD = \{(x_{\text{RGB}, i}, x_{\text{NIR}, i}, y_i) \, | \, i = 1,\dots, N\}
\end{equation}  
be the test data set, where the \(y_i\) are classification or segmentation
labels, and let \(\ell(f(x_{\text{RGB}}, x_{\text{NIR}}), y)\) be the relevant
accuracy function (0-1 loss for classification, Intersection-over-Union (IoU)
for segmentation). The test accuracy of \(f \) is then 
\begin{equation}
   \Acc (f, \cD) =\frac{1}{N} \sum_{i=1}^N \ell(f(x_{\text{RGB}, i}, x_{\text{NIR}, i}), y_i).
\end{equation}
To assess the importance of NIR information for model predictions, we use a
counterfactual dataset 
\begin{equation}
   \cD_{\text{NIR}, \sigma} = \{(x_{\text{RGB}, i}, x_{\text{NIR}, \sigma(i)}, y_i) \, | \, i = 1,\dots, N\}
\end{equation} 
obtained by shuffling the NIR ``column'' of \(\cD \) with a random permutation
\(\sigma \) of \(\{1,\dots,N\}\). In other words, the data points
\((x_{\text{RGB}, i}, x_{\text{NIR}, \sigma(i)}, y_i)\) consist of a labelled
RGB image \((x_{\text{RGB}, i},  y_i)\) together with the NIR channel
\(x_{\text{NIR}, \sigma(i)}\) of some other randomly selected data point in
\(\cD\). The \textbf{perceptual score of the NIR input is then}
\begin{equation}
   \label{eq:ps}
   \PS(f, \cD, \text{NIR}, \sigma) := \frac{\Acc (f, \cD)-\Acc (f, \cD_{\text{NIR}, \sigma})}{\Acc (f, \cD)}.
\end{equation}
In words, this is the relative accuracy drop incurred by evaluating \(f\) on the
dataset \(\cD_{\text{NIR}, \sigma}\) --- intuitively, if NIR input is important
to \(f\), replacing the NIR channel \(x_{\text{NIR}, i}\) with the NIR channel
\(x_{\text{NIR}, \sigma(i)}\) of some other randomly chosen image will
significantly damage the accuracy of \(f\), resulting in a large relative drop in
\cref{eq:ps}.

\begin{figure}[tb]
   \centering
   \includegraphics[width=\linewidth]{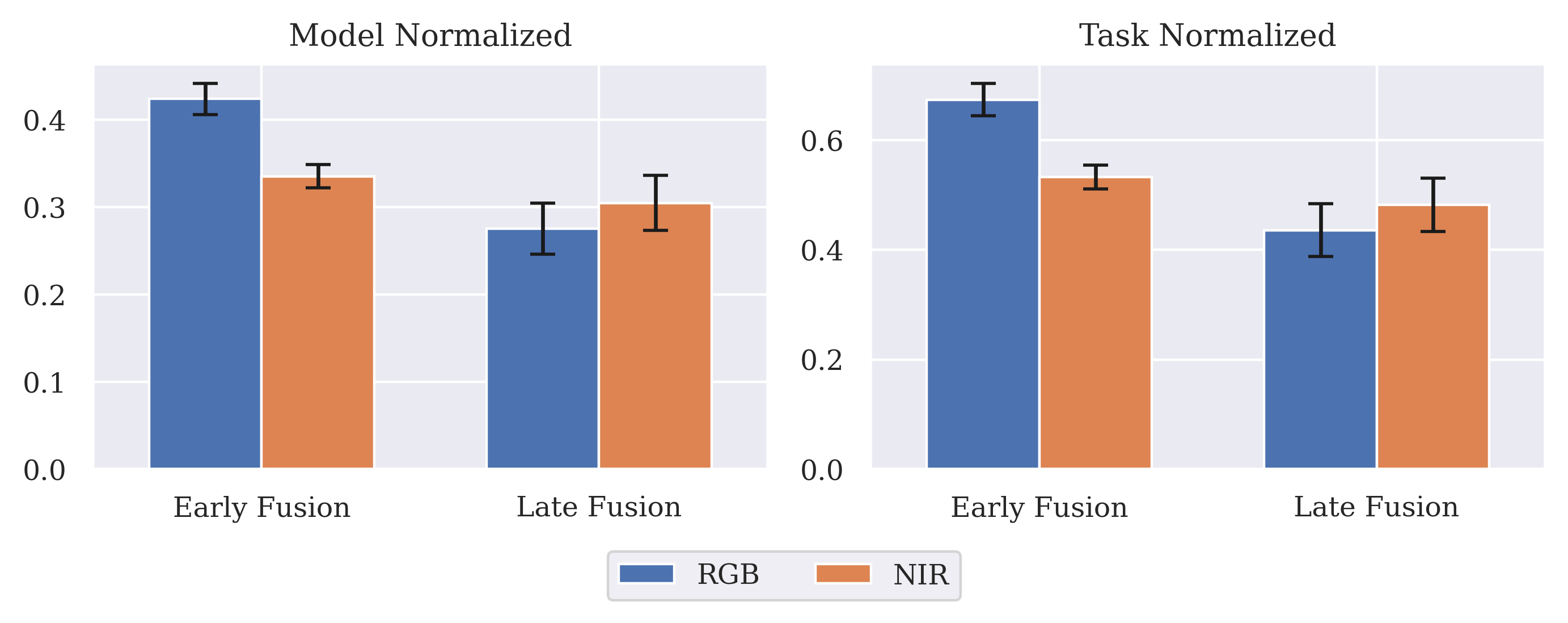}
   \caption{Perceptual scores for the RarePlanes multispectral classifiers. The
   early fusion models have a higher perceptual score for RGB channels (i.e., more reliance on RGB inputs), whereas
   the late fusion models have higher perceptual score for NIR channels (i.e., more reliance on NIR input). Error
   bars are obtained from five evaluations of the experiment with independent
   random number generator seeds.}\label{fig:rp-ps}
\end{figure}

The RGB perceptual score \(\PS(f, \cD, \text{NIR}, \sigma)\) is defined
analogously, permuting the RGB column instead of the NIR. In practice, we
average these metrics over several (e.g. 10) randomly selected permutations
\(\sigma \) of \(\{1,\dots,N\}\), and henceforth \(\sigma\) will be suppressed.
In fact \mycite{gat2021perceptual} defines two variants of perceptual score and
refers to the one in \cref{eq:ps} as ``model normalized''; their ``task
normalized'' variant uses majority vote accuracy (i.e. accuracy of a na\"ive
baseline) in the denominator instead of \(\Acc (f, \cD)\). We include both score normalizations for completeness, but note that in all cases the normalization did not change our qualitative conclusions. \Cref{fig:rp-ps}
displays these metrics for RarePlanes classifiers, and shows that from the
perspective of perceptual score, early fusion models pay more attention to
RGB channels whereas late fusion models pay more attention to NIR.
There is a simple heuristic explanation for these results: for both model
architectures, RGB channels occupy $75\%$ of the input space dimensions.
In contrast, in the late fusion models the RGB and NIR inputs are both encoded as
512-dimensional feature vectors which are concatenated before being passed to
the classification head (a 2 layer MLP); hence from the perspective of the
classification head, RGB and NIR each account for half of the feature
dimensions.

\begin{figure}[tb]
   \centering
   \includegraphics[width=\linewidth]{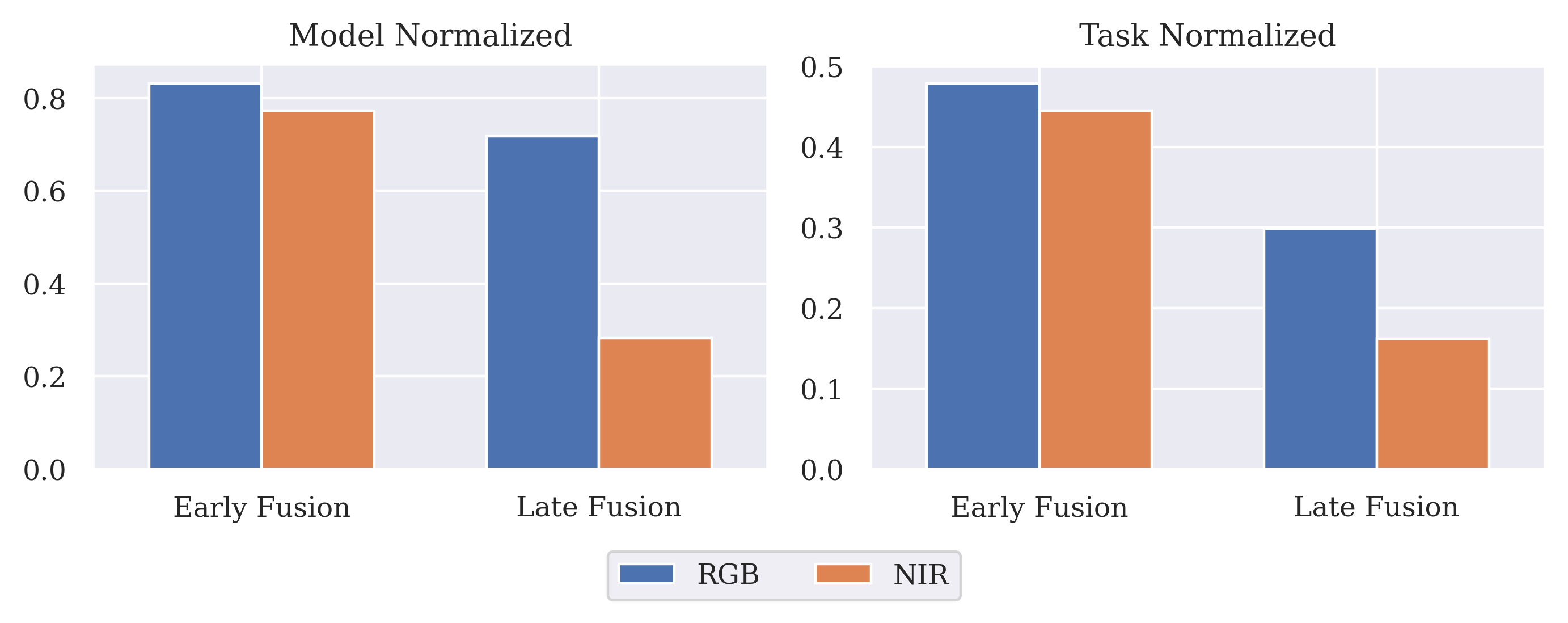}
   \caption{Perceptual scores for the US3D multispectral segmentation models. Both early and late fusion models have higher perceptual scores for RGB data, demonstrating that model performance relies more strongly on the RGB inputs. For late fusion models this effect is even more dramatic, suggesting that the NIR input is less important, in contrast to the classification model scores shown in \cref{fig:rp-ps}.}\label{fig:us3d-ps}
\end{figure}

Our measurements of perceptual score for segmentation models on US3D, shown in
\cref{fig:us3d-ps} are quite different: they suggest the late fusion model pays
even less attention (again, from the perspective of perceptual score) to NIR
information than the early fusion model. This finding is in tension with both
\cref{fig:rp-ps} and the heuristic explanation thereof. One challenge
encountered in interpreting \cref{fig:us3d-ps} is that due to the computational
cost of training we only trained one model for each architecture. We leave a
larger experiment allowing for estimation of statistical significance to future work.

\section{ROBUSTNESS TO NATURALISTIC CORRUPTIONS}
\label{sec:rob}

The perceptual scores presented in the previous section aim to quantify our
models' dependence on RGB and NIR inputs. A related question
is how robust these models are to naturally occurring corruptions that affect
either (or both) of the RGB or NIR inputs. With this in mind, we create corrupted
variants of RarePlanes and US3D by applying a suite of image transformations
simulating the effects of noise, blur, weather and digital corruptions. This is
accomplished using a fork of the code that generated
ImageNet-C,\mycite{hendrycksBenchmarkingNeuralNetwork2019} with modifications
allowing for larger images and more than 3 image channels. Each of the
corruptions applied comes with varying levels of severity (1 to 5). Where appropriate, we ensured that these corruptions were applied consistently between the RGB and NIR channels (e.g., snow is added to the same part of the image for all channels).
Visualizations of the corruptions considered for a sample RarePlanes chip can be
found in \cref{sec:corrupting}. 
\begin{figure}[tb]
   \centering
   \includegraphics[width=\linewidth]{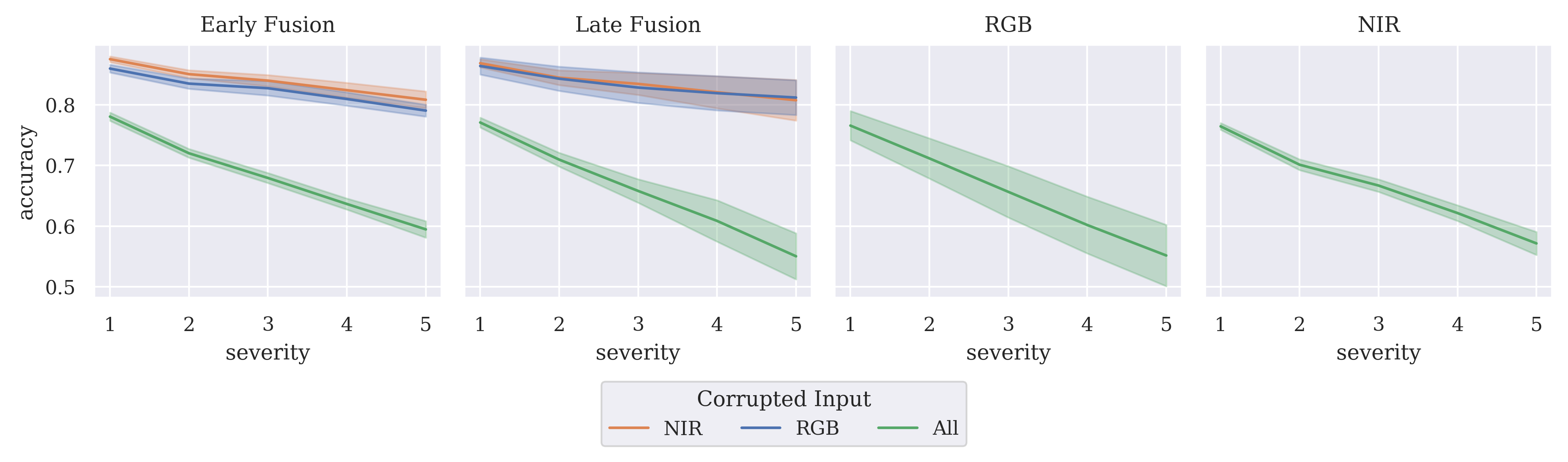}
   \caption{Corruption robustness of RarePlanes classifiers. Each subplot
   corresponds to a model architecture, and each line corresponds to a choice of
   input (RGB, NIR or both) to corrupt. Accuracy is averaged over all 15
   types of corruptions, and confidence intervals are obtained from five
   evaluations of the experiment with independent random number generator
   seeds.}\label{fig:rp-rob}
\end{figure}
We evaluate each model on the corrupted images. \Cref{fig:rp-rob} shows how accuracy of RarePlanes classifiers degrades with increasing corruption severity. We can see that when all channels in an image are corrupted, there is a similar drop in performance for all of the models considered (i.e., all four channels input to an early fusion model or all three channels input to an RGB model). This suggests that none of the architectures considered provides a significant increase in overall robustness to natural corruptions.

For the early and late fusion models, we also test model performance when either RGB or NIR (but not both) inputs are corrupted. For the late fusion model, the effects of corrupting one (but not both) of
\{RGB, NIR\} are more or less equal, while the early fusion model suffers a slightly greater drop in performance when RGB (but not NIR) channels are corrupted. We note that the confidence intervals in the early fusion model overlap; however, these results point in the same general direction as our perceptual score conclusions \cref{fig:rp-ps}: for early fusion models, RGB channels are weighted more heavily in predictions, and hence performance suffers more when RGB inputs are corrupted.

\begin{figure}[tb]
   \centering
   \includegraphics[width=\linewidth]{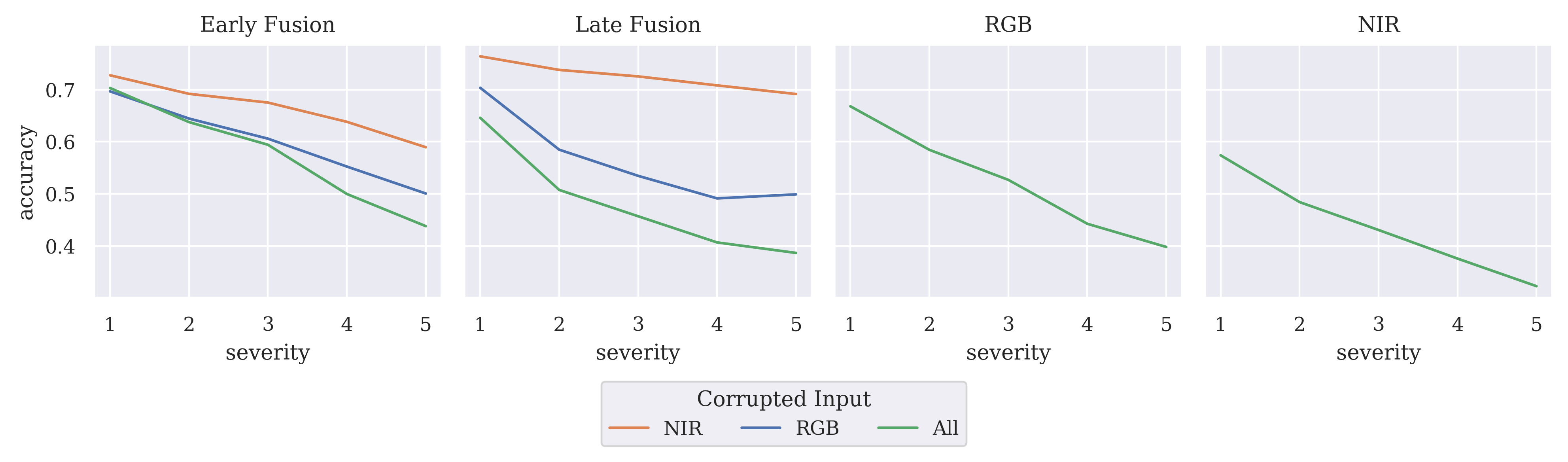}
   \caption{Corruption robustness of US3D segmentation models. Each subplot
   corresponds to a model architecture, and each line corresponds to a choice of
   input (RGB, NIR or both) to corrupt. IoU is averaged over all 15
   types of corruptions.}\label{fig:us3d-rob}
\end{figure}
\Cref{fig:us3d-rob} shows how performance of US3D segmentation models (as
measured by IoU) degrades with increasing corruption severity. When all input channels are corrupted (green line), the models show similar overall drops in performance, with the exception of the NIR-only model. The NIR-only model shows a larger drop in performance at all corruption severities, potentially suggesting that single-channel models are less robust to these kinds of natural corruptions.
For both early and late fusion architectures, a greater performance drop is incurred when RGB (but not NIR) channels are corrupted. In fact, corrupting only RGB channels is
almost as damaging as corrupting all inputs (both RGB+NIR). Notably in the case of late fusion, there is a large gap in robustness to corruptions of NIR inputs alone or RGB inputs alone. As was the case for the RarePlanes
experiments, these corruption robustness results point in the same general
direction as the perceptual score calculations in \cref{fig:us3d-ps}. For
example, the late fusion model had lower NIR perceptual scores than its
early fusion counterpart (i.e., less reliant on NIR inputs), and it is  more robust to NIR corruptions than the early fusion model.

We reiterate that due to computational costs we only train one US3D
segmentation model of each architecture, but with this caveat it does appear
that overall robustness in the case where all input channels are corrupted is
decreasing from left to right in \cref{fig:us3d-rob}. That is, the ranking of
models according to corruption robustness is: early, late, rgb, nir. One
potential explanation is that having more input channels (and hence more parameters) provides more
robustness, although this would not explain why early fusion models seem to be
more robust.\footnote{It has been found both theoretically and empirically that models with greater capacity in terms of number of parameters are \emph{capable} of greater robustness --- see for example \mycite{robustness_via_isoperim}.} Another possible explanation is that our models exhibit (positive) correlation between accuracy on clean test data and accuracy on corrupted data, as has been previously observed in the literature on robust RGB image classifiers \mycite{miller2021accuracy}. Indeed, in \cref{fig:us3d-iou} we see that the ranking of our US3D segmentation models by test IoU is: early, late (tied), rgb, nir. 

\section{LIMITATIONS AND OPEN QUESTIONS}

One limitation of our experiments is that the tasks considered are already
tractable by a deep learning model using only RGB images; incorporating
additional spectral bands offers at most incremental improvement. It would be
interesting to carry out the evaluations of \cref{sec:ps,sec:rob} for datasets
and models that more obviously benefit from additional multispectral information
beyond RGB. These might include ML models designed for tasks involving materials that can not be distinguished by RGB colors alone or environmental conditions in which NIR information is inherently valuable (such as pedestrian detection at night with RGB+NIR). 

In this work we considered two basic forms of multispectral fusion
(early and late), and although these arguably represent two interesting extremes
there are many more sophisticated architectural designs for fusing multiple
model
inputs \mycite{jayakumarMultiplicativeInteractionsWhere2020,baltrusaitisMultimodalMachineLearning2019,jayakumarMultiplicativeInteractionsWhere2020}.

Evaluating robustness of image segmentation models to naturalistic corruptions
is more complicated than in the case of classification tasks --- In particular,
there are some corruptions for which one might consider modifying segmentation
labels in parallel with the underlying images (one example is the ``elastic
transform'' corruption used in our experiments). In this work we did not apply
any modifications to segmentation labels. 

Finally, while we apply the same corruption algorithm to both the RGB and NIR
channels, in some cases this is not physically realistic, for example snow is in
fact \emph{dark} in infrared channels. For a study of test-time robustness to
more physically realistic corruptions of multispectral images, as well as
robustness (or lack thereof) of multispectral deep learning models to
adversarial corruptions of training data (i.e. data poisoning), we refer to
\mycite{bishoffQuantifyingRobustnessDeep2023}.

\section{CONCLUSION}

This work evaluates the extent to which multispectral fusion neural
networks with different underlying architectures 
\begin{enumerate}[(i)]
   \item pay differing amounts of
   attention to different input spectral bands (RGB and NIR) as measured by the
   perceptual score metric and
   \item exhibit varying levels of robustness to naturalistic corruptions
   affecting one or more input spectral bands.
\end{enumerate}
We find that the answers to (i) and (ii) correlate as one might expect: paying
more attention to RGB channels results in greater sensitivity to RGB
corruptions. Interestingly, our experimental results for segmentation models on
the US3D dataset contrast with those for classification models on the RarePlanes
datsets: In the classification experiments, early fusion models had higher
perceptual scores for RGB inputs, and late fusion models had slightly higher
perceptual scores for NIR inputs, whereas both types of fusion segmentation
models had higher perceptual scores for RGB inputs and the effect was more
extreme for late fusion (results for corruption robustness follow this trend).
This suggests that classification and segmentation models may make use of
multispectral information in quite different ways.

\appendix    

\section{IMAGE PROCESSING}
\label{sec:img-proc}


\subsection{RarePlanes}

The RarePlanes dataset includes both 8-bit RGB satellite imagery and 16-bit
8-band multispectral imagery, plus a panchromatic band. One of the goals of this
work is to assess the utility of including additional channels as input to image
segmentation models (e.g., near-infrared channels). In order to include channels
beyond Red, Green, or Blue, we must work from the 16-bit 8-band images. We
briefly describe our process for creating 8-bit, 8-band imagery, which consists
of pansharpening, rescaling, contrast stretching, and gamma correcting the
pixels in each channel independently. Specifically, the multispectral image is
pansharpened to the panchromatic band resolution using a weighted Brovey
algorithm \mycite{gillespie1987color}. The original 16-bit pixel values are
rescaled to 8-bit, and a gamma correction is applied using \(\gamma=2.2\)
\mycite{GammaCorrection2023}. The bottom 1\% of the pixel cumulative distribution
function is clipped, and the pixels are rescaled such that the minimum and
maximum pixel values are 0 and 255. 

We note that when applied to the R, G, and B channels of the multispectral image
products to generate 8-bit RGB images, this process produces images that are
visually similar but \emph{not} identical to the RGB images provided in Rare
Planes. As such, the RGB model presented in this work cannot be perfectly
compared to models published elsewhere trained on the RGB imagery included in
RarePlanes.\footnote{We trained identical models on the RarePlanes RGB images
and the RGB images produces in this work, and found that the RarePlanes models
performed negligibly better, at most a 1-2\% improvement in average accuracy.
This is likely due to more complex and robust techniques used for contrast
stretching and edge enhancement in RarePlanes; unfortunately these processing
pipelines are often proprietary and we could not find any published details of
the process.} However, we felt that this approach provided the most fair
comparison of model performance for different input channels, since the same
processing was applied identically to each channel.

\subsection{Urban Semantic 3D}

US3D builds upon the SpaceNet Challenge 4 dataset (hereafter SN4) \mycite{SN4}.
SN4 was originally designed for building footprint estimation in off-nadir
imagery, and includes satellite imagery from Atlanta, GA for view angles
between 7 an 50 degrees. US3D uses the subset of Atlanta, GA imagery from SN4
for which there exist matched LiDAR observations, and adds additional matched
satellite imagery and LiDAR data in Jacksonville, FL and Omaha, NE. The Atlanta
imagery is from Worldview-2, with ground sample distances (GSD) between 0.5m
and 0.7m, and view angles between 7 and 40 degrees. The Jacksonville and Omaha
imagery from Worldview-3, with GSD between 0.3m and 0.4m, and view angles
between 5 and 30 degrees. As described below, we train and evaluate models
using imagery from all three locations. We note however, that models trained
solely on imagery from a single location will show variation in overall
performance due to the variations in the scenery between locations (e.g.,
building density, seasonal changes in foliage and ground cover). 

US3D includes both 8-bit RGB satellite imagery and 16-bit pansharpened 8-band
multispectral imagery. Our process for creating 8-bit, 8-band imagery is similar
to the process we used for RarePlanes, the main exception being that we omit
pansharpening since it has already been applied to the multispectral images in
US3D. The original 16-bit pixel values are rescaled to 8-bit, and a gamma
correction is applied using $\gamma=2.2$. The bottom 1\% of the pixel cumulative
distribution function is clipped, and the pixels are rescaled such that the
minimum and maximum pixel values in each channel are 0 and 255. 

The US3D images are quite large (hundreds of thousands of pixels on a side) and
must be broken up into smaller images in order to be processed by a segmentation
deep learning model, a process sometimes called "tiling." Each of the large
satellite images (and matched labels) was divided into 1024 pixel x 1024 pixel
"tiles" without any overlap, producing 27,021 total images. All tiles from the
same parent satellite image are kept together during the generation of training
and validation splits to avoid cross contamination that could artificially
inflate accuracies\footnote{We note this is different from the data split
divisions within US3D, which mixes tiles from the same parent image between
training, validation, and testing.}. An iterative approach was used to divide
the satellite images into training, validation, and unseen hold-out (i.e., test)
splits to ensure that the distributions of certain meta-data properties
(location (Atlanta, Jacksonville, Omaha), view-angle, and azimuth angle) are
relatively similar from one split to the next; in particular, this avoids the
possibility that all images from a single location land in a single split. The
final data splits included 21,776 tiles in training (70\%), 2,102 tiles in
validation (8\%), and 3,142 tiles in the unseen, hold-out test split (12\%).
Models with near-infrared (NIR) input use the WorldView NIR2 channel, which
covers 860-1040nm. The NIR2 band is sensitive to vegetation but is less affected
by atmospheric absorption when compared with the NIR1 band. Segmentation labels
are stored as 8-bit unsigned integers between 0 and 255 in TIF files; during
training and evaluation we re-index these labels to integers between 0 and 6. We
retain the "unclassified" labels during model training and evaluation, but do
not include these classes in any metrics that average across all classes. 

As in the case of RarePlanes, when this process is applied to the R, G, and B
channels of the multispectral image products to generate 8-bit RGB images, it
produces images that are visually similar but \emph{not} identical to the RGB
images provided in US3D. As such, the RGB model presented in this work cannot be
perfectly compared to models published elsewhere trained on the RGB imagery
included in US3D.\footnote{We trained identical models on the US3D RGB images
and the RGB images produced in this work and found that the US3D models
performed slightly better (1-2\% improvement in average pixel accuracy). The
reasons for this improvement are likely similar to those described in the case
of RarePlanes (more complex and robust processing pipelines, the details of
which are unavailable).
} Again, we felt that
this approach provided the most fair comparison of model performance between
different input channels, since the same processing technique is applied
identically to each channel.

\subsection{Applying Corruptions}
\label{sec:corrupting}

\begin{figure}[tb]
   \centering
   \includegraphics[width=\linewidth]{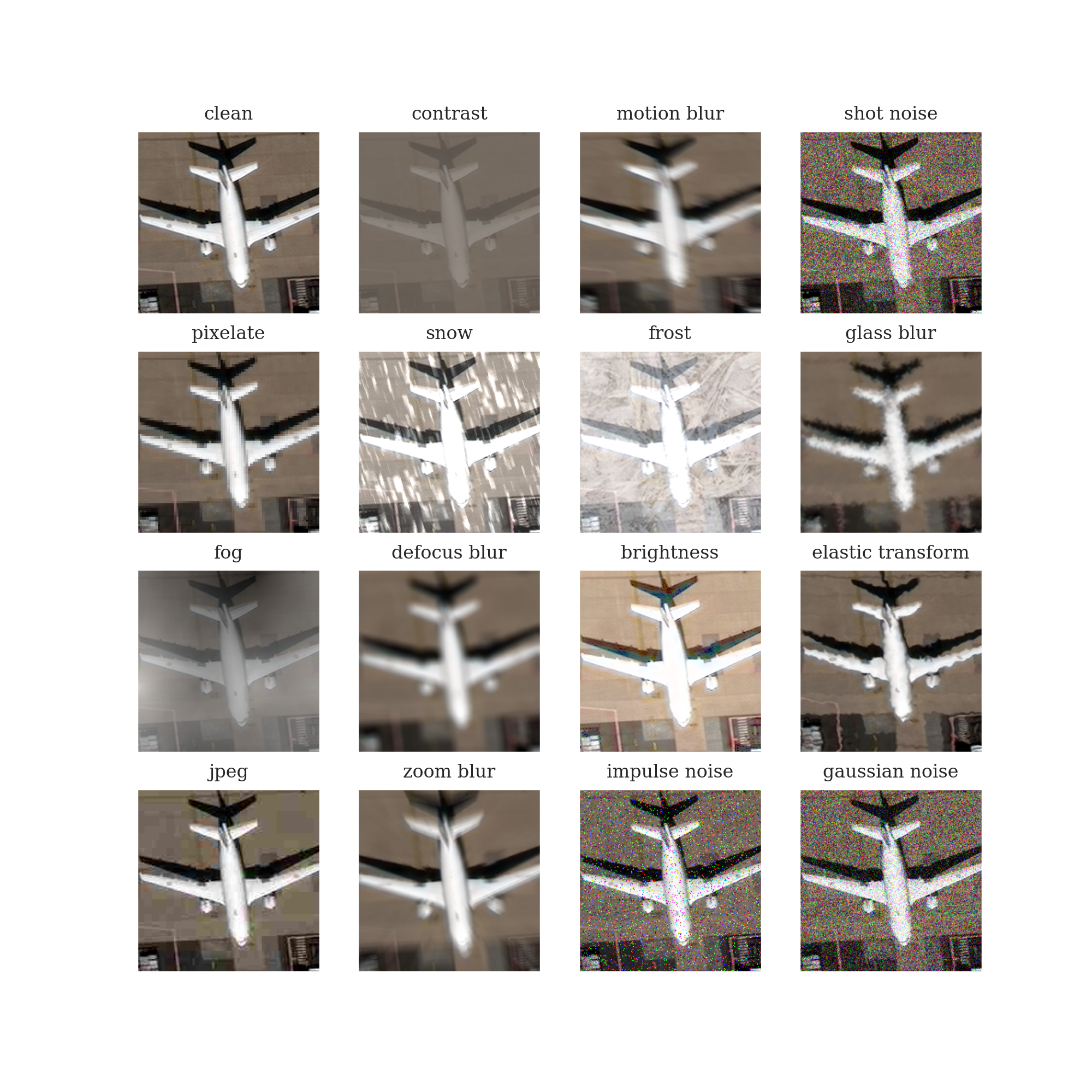}
   \caption{RGB corruptions of a RarePlane chip from our test set (severity level 3).}\label{fig:sample-corr-rgb}
\end{figure}
\begin{figure}[tb]
   \centering
   \includegraphics[width=\linewidth]{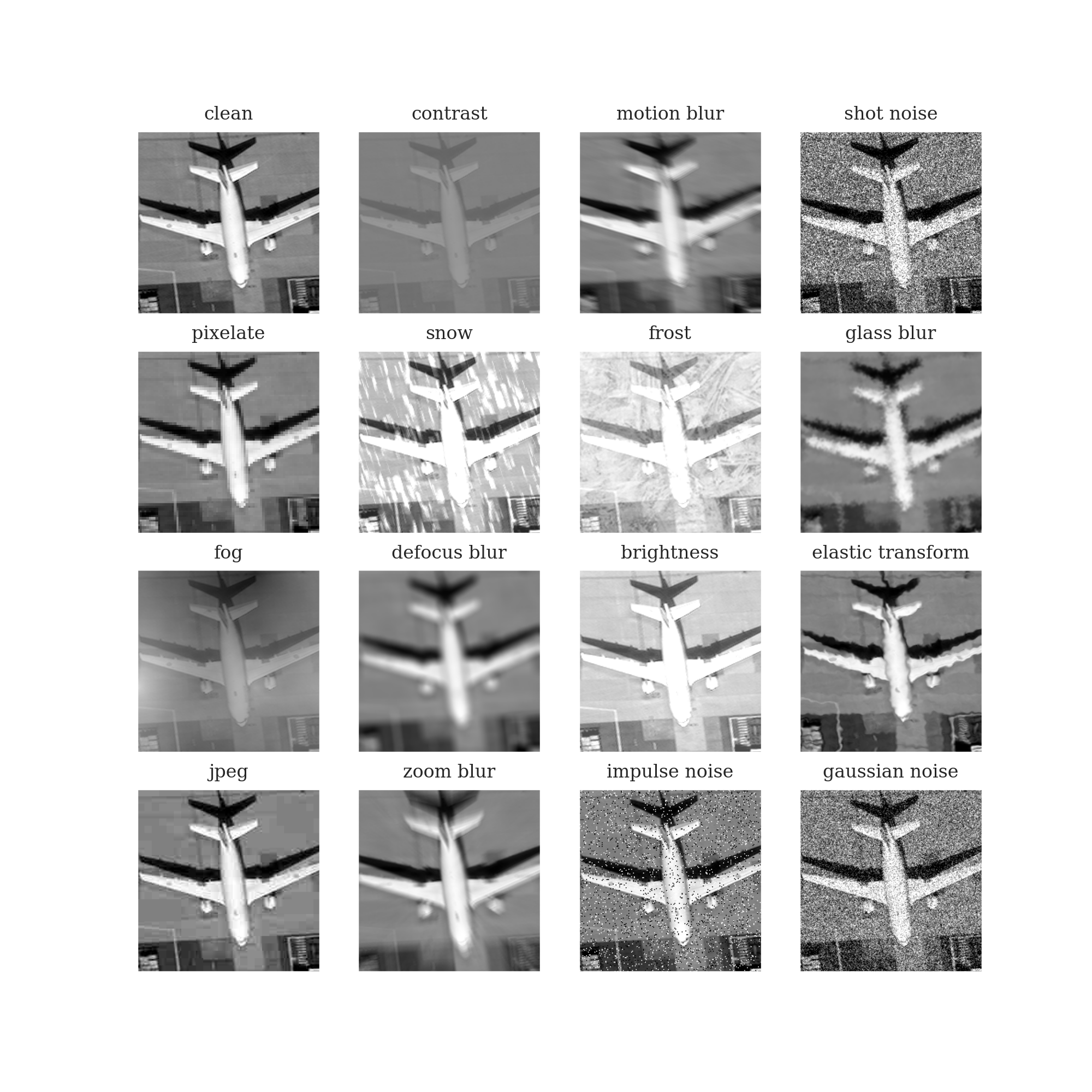}
   \caption{NIR corruptions of a RarePlane chip from our test set (severity level 3). Note that the motion blur (2nd row, 3rd column) is applied in the same direction as in \cref{fig:sample-corr-rgb}.}\label{fig:sample-corr-nir}
\end{figure}

As mentioned above, we modify the code available at
\href{https://github.com/hendrycks/robustness}{github.com/hendrycks/robustness}
(hereafter referred to as ``the \texttt{robustness} library'' or simply ``\texttt{robustness}''),
which was originally designed for \(224\times 224\) RGB images, to achieve the
following goals:
\begin{enumerate}[(i)]
   \item Arbitrary image resolution and aspect ratio (in fact this feature was
   already implemented in \mycite{michaelis2019dragon}, though we did not
   discover that repository until after this work was completed). This was
   essential as the RarePlanes ``chips'' have varying resolution and aspect
   ratio, and while all US3D tiles are of shape \(1024\times 1024\), that
   differs from the \(224\times 224\) shape considered in \texttt{robustness}.
   \item Input channels beyond RGB. 
\end{enumerate}
When applying corruptions to RGB+NIR images, we separate the 4-channel image
into two 3-channel images, one containing the RGB channels and the other
consisting of three stacked copies of the NIR channel. We then apply corruptions
from \texttt{robustness} separately to each of these 3-channel images, with the
following consideration: wherever physically sensible, we use the same
randomness for both the RGB and NIR input. For example in the case of motion
blur, we use the same velocity vector for both --- not doing so would correspond
to a physically unrealistic situation where the RGB and NIR sensors are moving
in independent directions. On the other hand for corruptions such as shot noise
modeling random processes affecting each pixel independently we use independent
randomness for RGB and NIR. Corruptions of a RarePlanes test image can be seen
in \cref{fig:sample-corr-rgb,fig:sample-corr-nir}.

\begin{figure}[tb]
   \centering
   \includegraphics[width=0.9\linewidth]{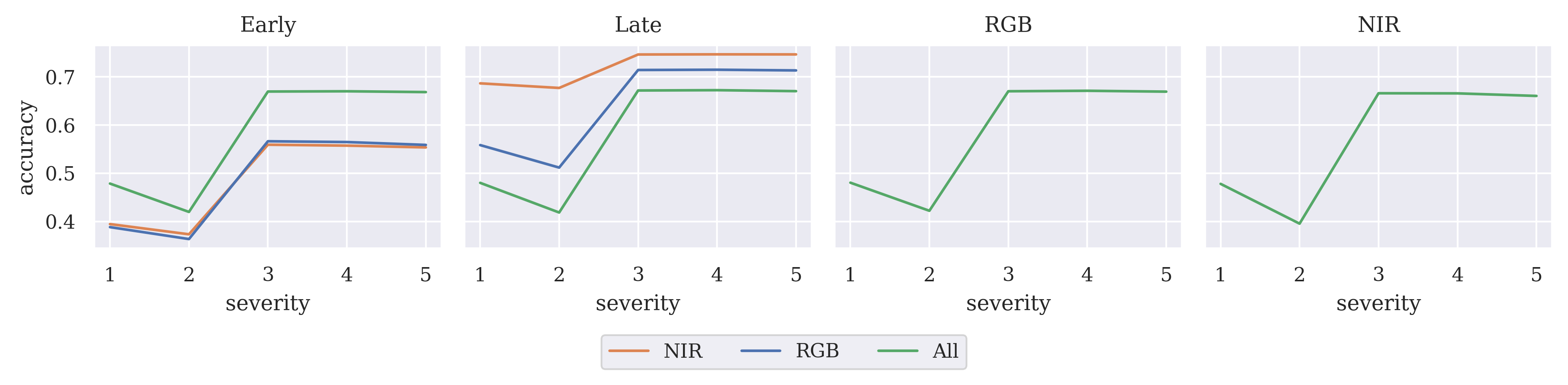}
   \caption{Results from the experiments summarized in \cref{fig:us3d-rob},
   restricting attention to the elastic transform
   corruption of \mycite{hendrycksBenchmarkingNeuralNetwork2019}.}
   \label{fig:et}
\end{figure}

We do not modify labels in any way. In the case of US3D one could argue that for
some corruptions the segmentation labels should be modified. For example,
``elastic transform'' is implemented by applying localized permutations of some
pixels and then blurring. Here it would make sense to apply at the exact same
localized permutations to the per-pixel segmentation labels and then possibly
blur them using soft labels (where each pixel is assigned to a probability
distribution over segmentation classes). One could also imagine using soft label
blurring for e.g. zoom or motion blur. Our reason for leaving segmentation
labels fixed is pragmatic: in most cases we found where an argument could be
made for modifying labels, doing so seemed to require working with \(\text{num.
classes} \times H \times W\) soft label tensors (possibly at an intermediate
stage), and this would require modifications to multiple components of our
pipeline, which (in keeping with standard practice) stores and utilizes labels
as 8-bit RGB images with segmentation classes encoded as certain colors.
Moreover, we emphasize that for most corruptions most severity levels the unmodified labels fairly reflect ground truth. With
all of that said, in the case of elastic transform mentioned above we observed
the bizarre experimental accuracy increasing as corruption severity increased
--- see \cref{fig:et}. Determining whether the results in that figure persist
even after more care is taken with segmentation labels is a high priority item
for future work.

\section{MODEL ARCHITECTURE AND TRAINING}
\label{sec:mod-arch-train}

For the RarePlanes experiments, all models are based on the ResNet34
architecture \mycite{heDeepResidualLearning2015}, pretrained on ImageNet (we use
the weights from \mycite{torchvision}). 

For the early fusion model, we stack the RGB and NIR input channels to obtain a
\(4\times H \times W\) input tensor, and replace the ResNet34 first layer
convolution weight, originally of shape \(3 \times C \times H \times W\), with a
weight of shape \(4 \times C \times H \times W\) by repeating the red channel
twice (i.e., we initialize the convolution weights applied to the NIR channel
with those that were applied to the red channel in the ImageNet pretrained
ResNet). For the pure NIR channel, we adopt a similar initialization strategy,
but discarding the RGB weights since we only use one-channel NIR inputs (so in
this case the first convolution weight has shape \(1 \times C \times H \times
W\)). For the pure RGB model of course no modifications are required.

In the late fusion model, we take a pure RGB and pure NIR architecture as
described above, and remove their final fully connected layers: call these
\(f_{\text{rgb}}\) and \(f_{\text{nir}}\). Given an input of the form
\((x_{\text{rgb}}, x_{\text{nir}})\), we use \(f_{\text{rgb}}\) and
\(f_{\text{nir}}\) to compute two 512-dimensional feature vectors
\(f_{\text{rgb}}(x_{\text{rgb}})\) and \(f_{\text{nir}}(x_{\text{nir}})\). These
are then concatenated to obtain a 1024-dimensional feature vector
\(\mathrm{cat}(f_{\text{rgb}}(x_{\text{rgb}}),
f_{\text{nir}}(x_{\text{nir}}))\). Finally, this is passed through a multi-layer
perceptron with two 512-dimensional hidden layers.

For all training and evaluation, we ``pixel normalize'' input images,
subtracting the a mean RGB+NIR
four-dimensional vector, and dividing by a corresponding standard deviation. In the RarePlanes experiments we use the mean and standard deviation of the pretraining dataset (ImageNet) for RGB channels, and a mean and standard deviation computed from our RarePlanes images for the NIR channel. 
We fine tune on RarePlanes for 100 epochs using stochastic gradient descent with initial
learning rate \(10^{-3}\), weight decay \(10^{-4}\) and momentum 0.9. We use
distributed data parallel training with effective batch size 256 (128 \(\times\)
2 GPUs). We use a ``reduce-on-plateau'' learning rate schedule that multiplies
the learning rate by \(0.1\) if training proceeds for 10 epochs without a
\(1\%\) increase in validation accuracy. We train five models of each fusion
architecture with independent random seeds (randomness in play includes new
model layers (all models include at least a new final classification layers) and
SGD batching).   

\begin{figure}[tb]
   \centering
   \begin{subfigure}{0.48\linewidth}
      \centering
      \includegraphics[width=\linewidth]{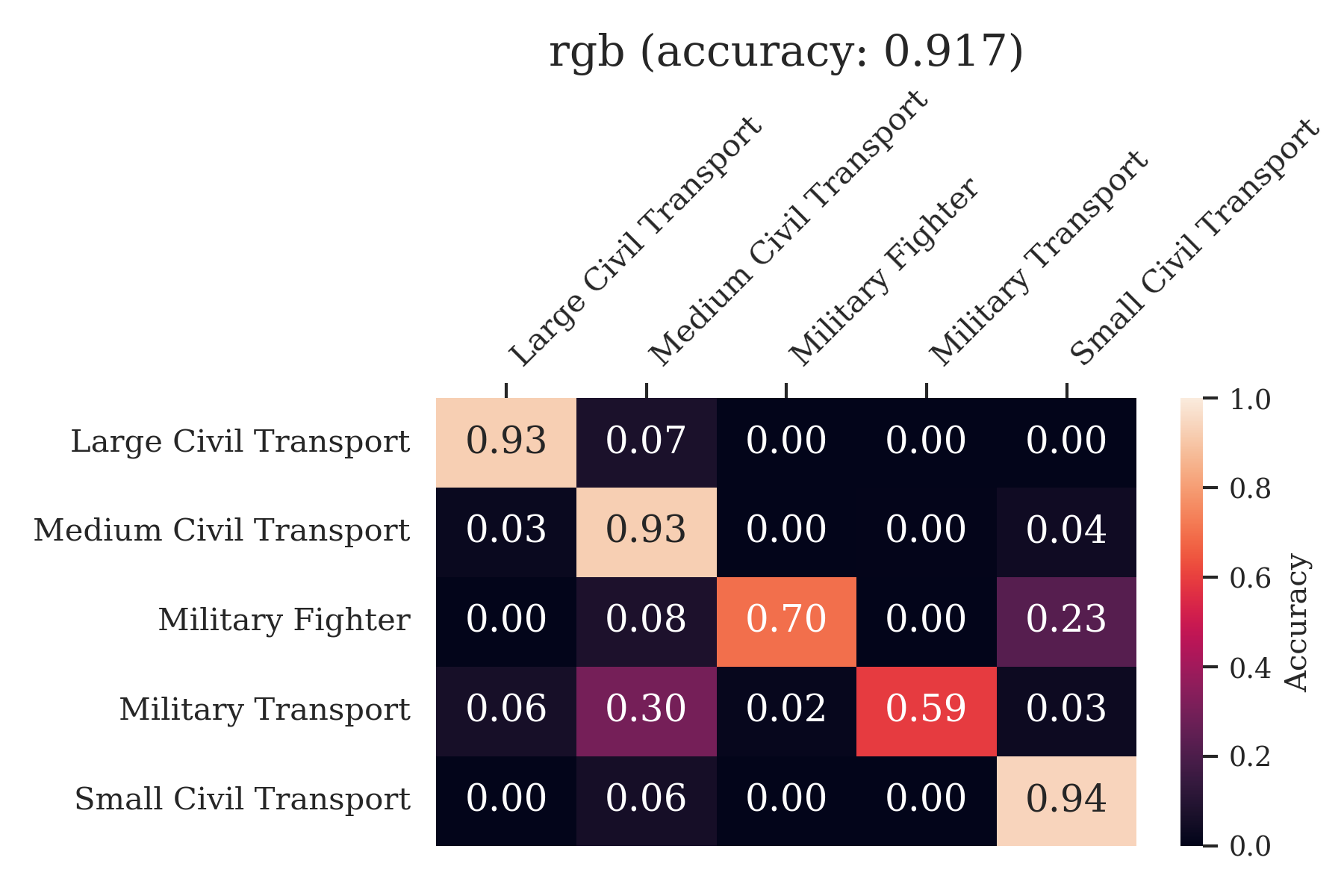}
   \end{subfigure}
   \begin{subfigure}{0.48\linewidth}
      \centering
      \includegraphics[width=\linewidth]{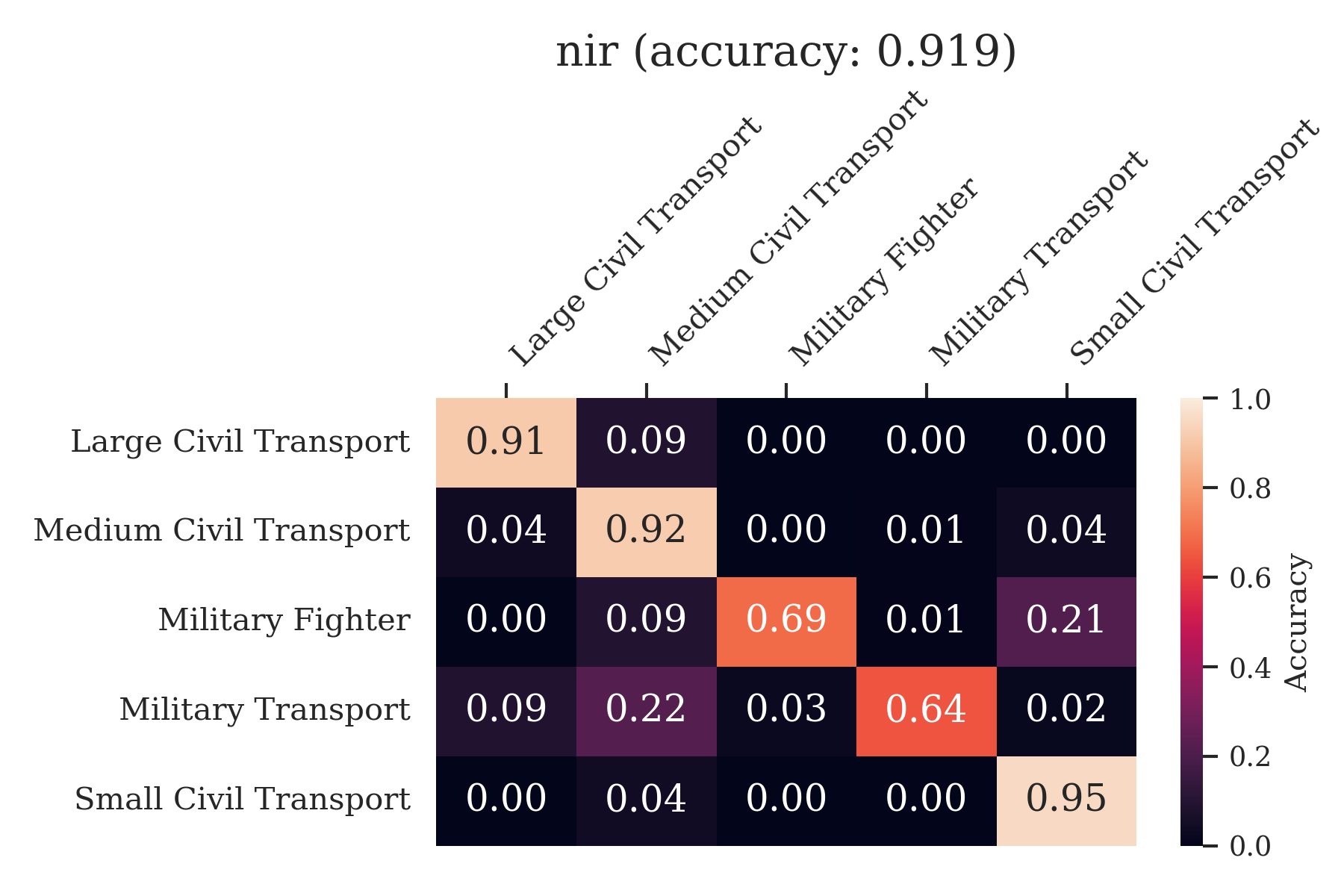}
   \end{subfigure}
   \begin{subfigure}{0.48\linewidth}
      \centering
      \includegraphics[width=\linewidth]{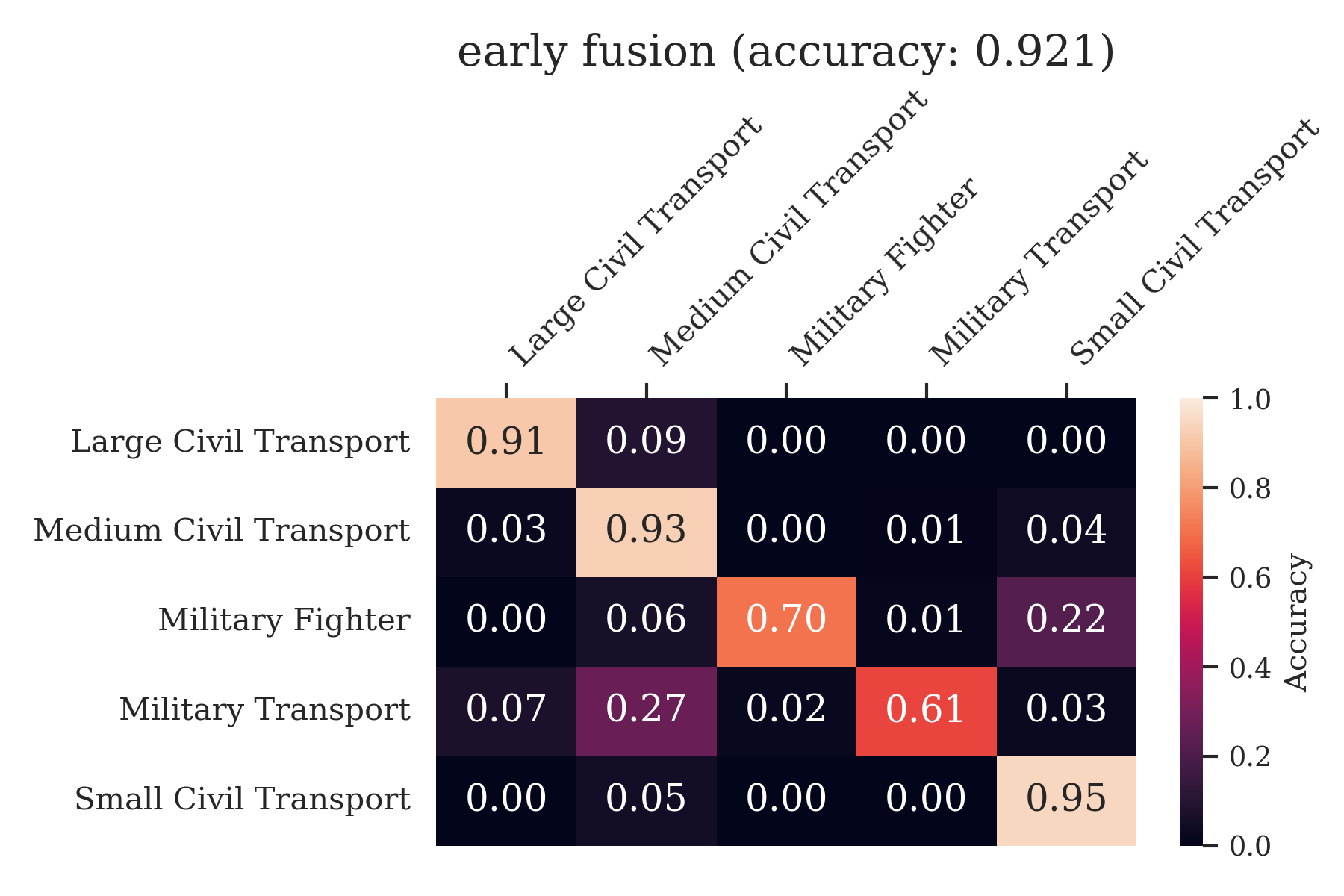}
   \end{subfigure}
   \begin{subfigure}{0.48\linewidth}
      \centering
      \includegraphics[width=\linewidth]{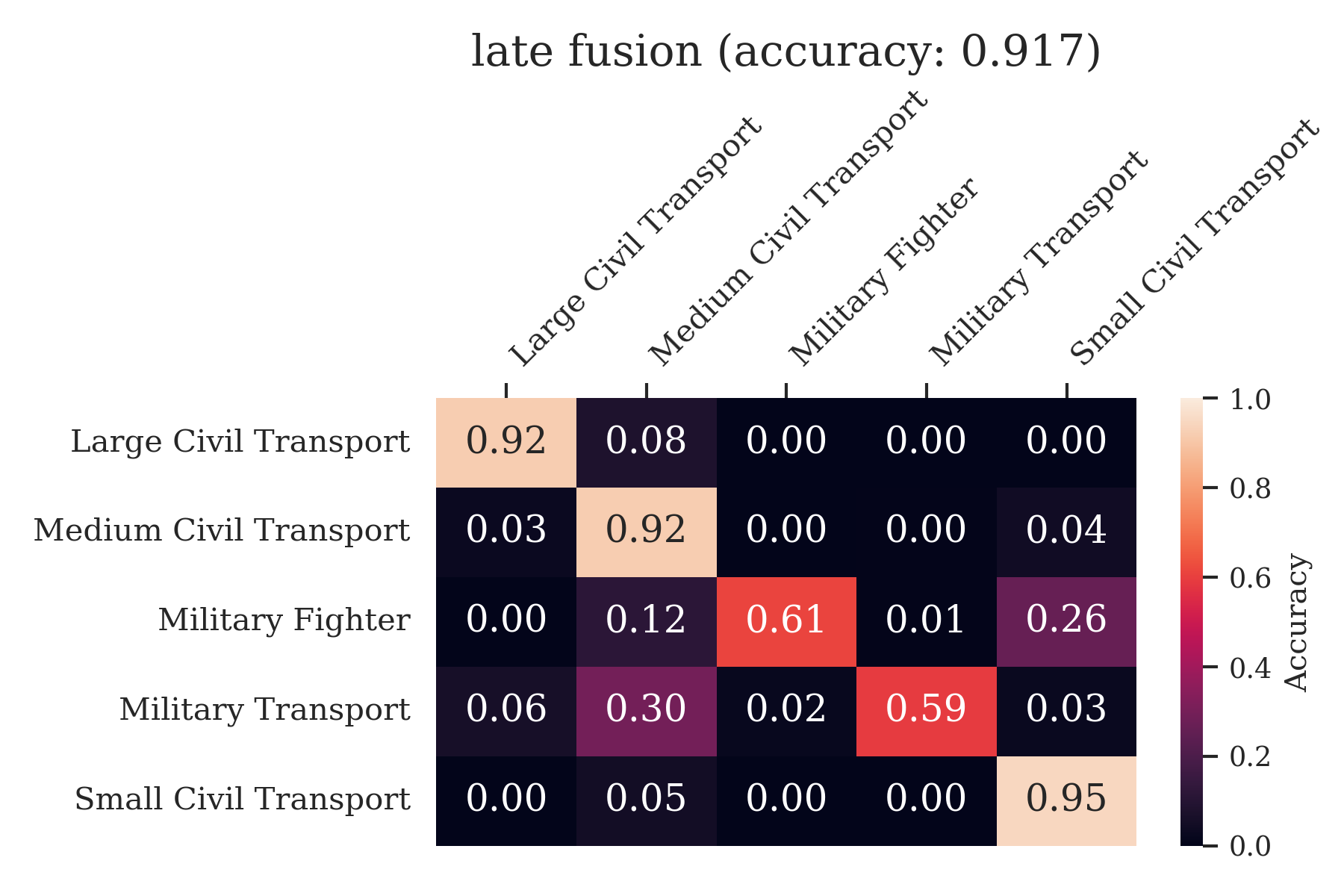}
   \end{subfigure}
   \caption{Test accuracies and confusion matrices for RGB+NIR RarePlanes models.}
   \label{fig:rp-conf}
\end{figure}

\Cref{fig:rp-conf} displays test accuracies and class confusion matrices of our
trained RarePlanes classifiers. The accuracies of \(\approx 92 \%\) are
respectable but by no means state-of-the-art.

All of our US3D segmentation models are based on DeepLabv3 with a ResNet50 backbone,
pretrained on COCO \mycite{linMicrosoftCOCOCommon2014} (again obtained from
\mycite{torchvision}). Our methods for defining RGB+NIR fusion architectures are
similar to those described above for image classifiers, the difference being
that we modify the ResNet50 backbone. In the case of early fusion and pure NIR
models, we only need to modify the first layer convolution weights as described
for the RarePlanes classifiers (and of course for the pure RGB model no
modifications are necessary). 

For the late fusion segmentation model, we start with pure RGB and pure NIR ResNet50
backbones: call these
\(f_{\text{rgb}}\) and \(f_{\text{nir}}\). Given an input of the form
\((x_{\text{rgb}}, x_{\text{nir}})\), we use \(f_{\text{rgb}}\) and
\(f_{\text{nir}}\) to compute two 2048-dimensional feature vectors
\(f_{\text{rgb}}(x_{\text{rgb}})\) and \(f_{\text{nir}}(x_{\text{nir}})\). These
are then concatenated to obtain a 4096-dimensional feature vector
\(\mathrm{cat}(f_{\text{rgb}}(x_{\text{rgb}}),
f_{\text{nir}}(x_{\text{nir}}))\) which is then passed through the DeepLabv3
atrous convolution segmentation ``head.'' 

In the US3D experiments we use the mean and standard deviation of ImageNet for RGB channels, and a mean and standard deviation of the ImageNet R channel for the NIR channel.\footnote{Note that while Torchvision's DeepLabv3 was pretrained on COCO, not ImageNet, inspection of their preprocessing (\texttt{tvmodsseg.DeepLabV3\_ResNet50\_Weights.DEFAULT.transforms}) shows that the \emph{ImageNet} mean and standard deviation were used for normalization!} 
We fine tune on US3D for 400 epochs using the Dice loss function optimized with
Adam \mycite{adam} with initial learning rate \(5 \times 10^{-4}\) and weight
decay \(10^{-5}\) (and PyTorch \mycite{pytorch} defaults for all other Adam
hyperparameters). We use distributed data parallel training with effective batch
size 32 (4 \(\times\) 8 GPUs). We use a ``reduce-on-plateau'' learning rate
schedule that multiplies the learning rate by \(0.5\) if training proceeds for
25 epochs without a relative \(1\%\) increase in validation accuracy. 

\begin{figure}[tb]
   \centering
   \includegraphics[width=0.7\linewidth]{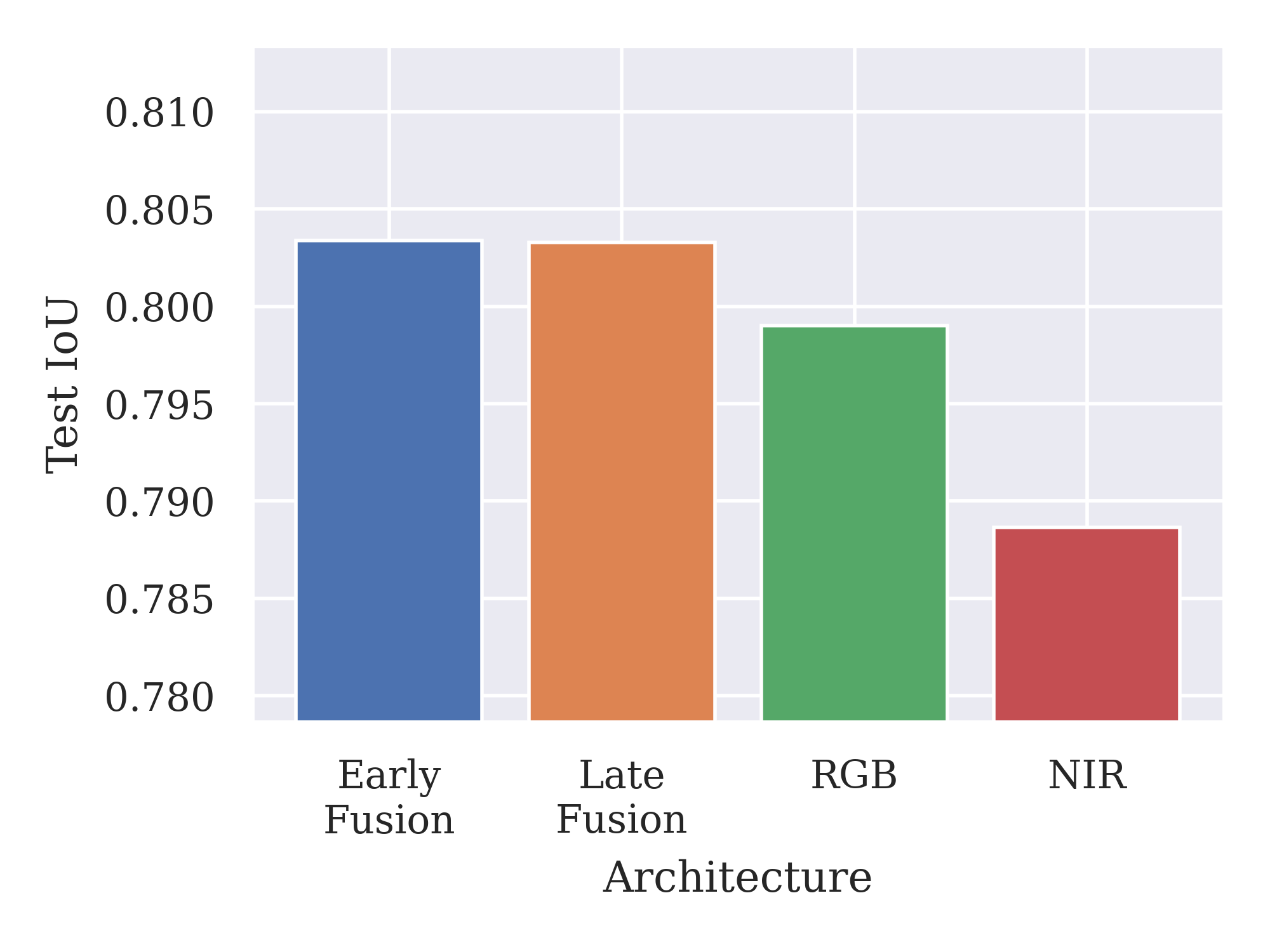}
   \caption{Intersection-over-Union (IoU) of US3D segmentation models.}
   \label{fig:us3d-iou}
\end{figure}

\Cref{fig:us3d-iou} displays test Intersection-over-Union (IoU) of our
trained US3D segmentation models.

\section*{ACKNOWLEDGMENTS}       

The research described in this paper was conducted
under the Laboratory Directed Research and Development Program at Pacific
Northwest National Laboratory, a multiprogram national laboratory operated by
Battelle for the U.S. Department of Energy.

\clearpage
\bibliographystyle{spiebib} 
\bibliography{arch-impact} 

\end{document}